\documentclass{article}
\usepackage[utf8]{inputenc}
\usepackage{microtype}
\usepackage{graphicx}

\usepackage{subcaption}
\usepackage{booktabs} %

\usepackage{xcolor}
\definecolor{orange}{HTML}{fd9d00} %
\definecolor{green}{HTML}{77fa00} %
\usepackage{amssymb}

\usepackage{hyperref}

\usepackage{amsmath}
\usepackage[capitalise]{cleveref}
\DeclareMathOperator{\Tr}{Tr}

\usepackage[accepted]{icml2020}

\icmltitlerunning{Client Adaptation improves FL with Simulated Non-IID Clients}

\begin{document}

\twocolumn[

\icmltitle{Client Adaptation improves Federated Learning\\ with Simulated Non-IID Clients}

\icmlsetsymbol{equal}{*}

\begin{icmlauthorlist}
\icmlauthor{L.~Rieger}{equal,dtu}
\icmlauthor{R.~M.~Th.~H\o egh}{equal,dtu}
\icmlauthor{L.~K.~Hansen}{dtu}
\end{icmlauthorlist}

\icmlaffiliation{dtu}{DTU Compute, Technical University Denmark, 2800 Kgs. Lyngby, Denmark}

\icmlcorrespondingauthor{L.~Rieger}{lauri@dtu.dk}
\icmlcorrespondingauthor{R.~M.~Th.~H\o egh}{rmth@dtu.dk}

\icmlkeywords{federated learning, audio representation, distributed}

\vskip 0.3in
]

\printAffiliationsAndNotice{\icmlEqualContribution} %

\begin{abstract}
We present a federated learning approach for learning a client adaptable, robust model when data is non-identically and non-independently distributed (non-IID) across clients.
By simulating heterogeneous clients, we show that adding learned client-specific conditioning improves model performance, and the approach is shown to work on balanced and imbalanced data set from both audio and image domains.
The client adaptation is implemented by a conditional gated activation
unit and is particularly beneficial when there are large differences between the data distribution for each client, a common scenario in federated learning.

\end{abstract}

\section{Introduction}
As of 2019, an estimated three billion mobile phones are connected to the internet, collectively amassing a staggering amount of information \cite{Lim2019}, and it is highly desirable to make this data available for ever more data-demanding neural networks.
Traditional approaches for training neural networks require that all data is collected in one place for training.
However, as the data is often sensitive, and contains a wealth of private information about the user, centralized data collection is not always realizable or desirable.

Federated learning (FL) provides an approach to learn a centralized model from decentralized data in a privacy-aware manner \citep{McMahan2016}.
Local copies of a global model are trained at participating decentralized nodes (or clients) with local data, and the model is then consolidated episodically.
While there are privacy concerns about potential data leakage through model updates, federated learning is often preferable to training a model in a centralized manner \citep{9069945}.
Learning in this decentralized manner does, however, pose new problems for training neural networks, especially when the data available at individual clients is highly heterogeneous as well as sensitive from a privacy perspective.
An example of such highly sensitive data is audio collected by personal mobile devices.

A federated learning approach where the data does not leave the person's phone is particularly well-suited for this scenario.
Yet, such data from smartphones is challenging in multiple points.  Not only might the data be class imbalanced, but even conditioned on the class the feature distributions between clients can diverge, as people are in different environments with different soundscapes and use different phones to record the data.
\citet{Li2019a} noted that while the original federated learning algorithm FederatedAveraging algorithm (\textsc{FedAvg}) will converge under certain strong assumptions, it lacks this guarantee under this more realistic assumption that data distribution between clients will be diverse.

However, certain features of the input, while variant across clients, are likely to be consistent for a single client, as people tend to e.g. use the same phone for a long amount of time and do not change their voice or speaking patterns substantially.
Motivated by this, we propose to learn a local embedding for each client along with the global model as shown in \cref{fig:CGAU}.
We evaluate this approach, which we coined the conditional gated activation unit (CGAU), on two classification tasks, one from the audio domain and one from the image domain, covering both balanced and imbalanced data. CGAU outperforms a baseline in both scenarios, showing the usefulness of learning localized features explicitly as opposed to encoding them in the global model.

The current paper presents two contributions. Firstly, we propose the conditional gated activation unit (CGAU), an enhancement for current neural network architectures suited for federated learning that captures features in client-dependent non-IID data. CGAU can be utilized in conjunction with currently used federated learning algorithms such as \textsc{FedAvg}.  Secondly, to evaluate CGAU, we present a principled approach to simulate clients with non-IID data for evaluating federated learning. The approach utilizes embeddings from pre-trained networks to simulate clients by finding clusters in the embedding space.

\section{Background}
In 2016 \citet{mcmahan2016communication} coined the term federated learning to refer to the training of a model with data that is only available at many distributed devices. This setting is characterized by a few properties that challenge traditional machine learning approaches. For one,  the data is typically distributed on a large number of clients. Additionally, since there is a large variance in the behaviour and environment of the typical phone user, it is non-IID. Finally, the data is likely to be unbalanced, i.e. there is a large variance in the number of samples and class distributions at each client.

To combat those issues they proposed the \textsc{FedAvg}, a variant on traditional stochastic gradient descent (SGD).
\textsc{FedAvg} consolidates training updates from a large number of different sources with potentially unbalanced training data. Each client has a local copy of the model to be trained. In each round, a fraction of the total clients available computes the weight update of the model on their locally available data. The number of weight updates before global consolidation $E$ is a hyperparameter.
After $E$ weight updates on the local model, the client sends the local model to the server. The server averages the model weights from all clients into the global model.
By only synchronizing the local models every $E$ steps, the communication cost is greatly reduced, making \textsc{FedAvg} feasible for training on distributed clients.

Since \citet{mcmahan2016communication} a number of papers have followed up on this approach, aiming to e.g. reduce the necessary communication between clients, examine the vulnerability of the model towards adversarial attacks or deal with non-IID data \cite{konevcny2016federated,Sattler2019,Li2019a,bagdasaryan2018backdoor,nguyen2020poisoning}.
We focus specifically on approaches dealing with non-IID data, and we will in the following assume a traditional supervised learning task, learning a function that maps an input to an output class. This covers the most common tasks where federated learning is used.

\citet{Zhao2018} showed that federated learning is vulnerable to data being non-IID in the distribution of classes with the extreme case being that each client only sees one class.
They show that the difficulty arises because the model weights from different classes diverge too much before global synchronization. They also show how weight divergence in between clients is bounded by the earth mover's distance in between the distributions of classes from the clients and the global distribution. The model performance can be partly recovered by sharing a small proportion of the data globally to all clients. In contrast to our work, they only consider differences in distribution over the classes, not in characteristics of the input data (non-IID feature distributions as opposed to class distributions).

\citet{Sattler2019} proposed a new compression scheme, sparse ternary compression (STC) to reduce the communication necessary in a non-IID setting regarding the class distributions from the individual clients.

Recently \citet{Li2019a} further examined the influence of non-IID data on model performance. They show empirically and theoretically that heterogeneous data will slow down convergence of the model to the minimum and established that with non-IID data, the learning rate must be decayed over time for the model to converge to the optimal state.

\citet{Peng2019b} aim to mitigate the effect of domain shift heterogeneity by learning invariant features with adversarial reconstruction. They split up the embedding of the input into a domain-specific and domain-invariant part by minimizing mutual information between the two components. Then only the domain-invariant component is used for classification of the original task while an additional loss function is placed on a complete reconstruction of the input embedding from both. In this way, the network learns to disentangle the domain-invariant features and becomes more robust towards domain shift.

Finally, \citet{Ghosh2019} proposes a solution for training with heterogeneous data distributions on the clients. They propose first finding independent locally optimal solutions for each client. The clients then send their model to the server. The server clusters the clients based on the locally optimal solutions. Consequently, a traditional FL optimization is run for each client cluster.
Like in our work, they explicitly consider diversity in the data distributions.
However, in contrast to our work \citet{Ghosh2019} do not work with neural networks as it assumes a (relatively) low-dimensional representation of the learnt algorithm parameters.

\section{Methods}

\subsection{Federated learning with pre-trained networks}\label{sec:pre-trained}
Deep learning requires, in general, a large amount of data and computational power for training, and the resulting models are often of considerable size in terms of memory.
In settings where resources like computational power, data, and data transfer are limited, e.g., for federated learning on mobile devices, the resource demands of training deep neural networks from scratch can be prohibitively large. Pre-trained neural networks, fine-tuned on the task at hand via transfer-learning present a viable solution for these issues.

We investigate the use of pre-trained networks for classifications tasks in a federated learning scheme, where a pre-trained network is used as a ``frozen'' feature extractor (that is, the pre-trained network is not further trained using federated learning).
By using a pre-trained network, we off-load the needed computational power and data required from the federated learning process as the feature extractor network can be trained centrally using any available but task-relevant data, or taken from already trained networks from a relevant domain.
Additionally, the communication costs are reduced, as the feature extractor does not need to be sent back and forth between rounds of federated learning.
The federated learning then only has to learn to solve the problem of interest by learning a (much smaller) classifier on top of the embedding that the pre-trained network produces.

\subsection{Client adaptation through conditional gated activation units} \label{sec:cgau}
Heterogeneous clients with features that are non-IID are an inevitable part of federated learning in realistic settings.
Given a set of completely homogeneous clients, anything a model learns based on a particular client would generalize to other clients.
With heterogeneous clients, however, some class characteristics for a particular client might, or might not, generalize to other clients.
Similarly, some shared class characteristics might be expressed differently at each client.
Learning in such heterogeneous settings is more difficult, in part, as the information that can be shared between clients is reduced (e.g. by patterns distinct to a subset of clients, which we will call client-specific expression) and the information that can be shared is obfuscated (the same underlying pattern looks somewhat different at different clients, which we will call client-specific modulation).

\begin{figure}
 \centering
 \includegraphics[width=.35\textwidth]{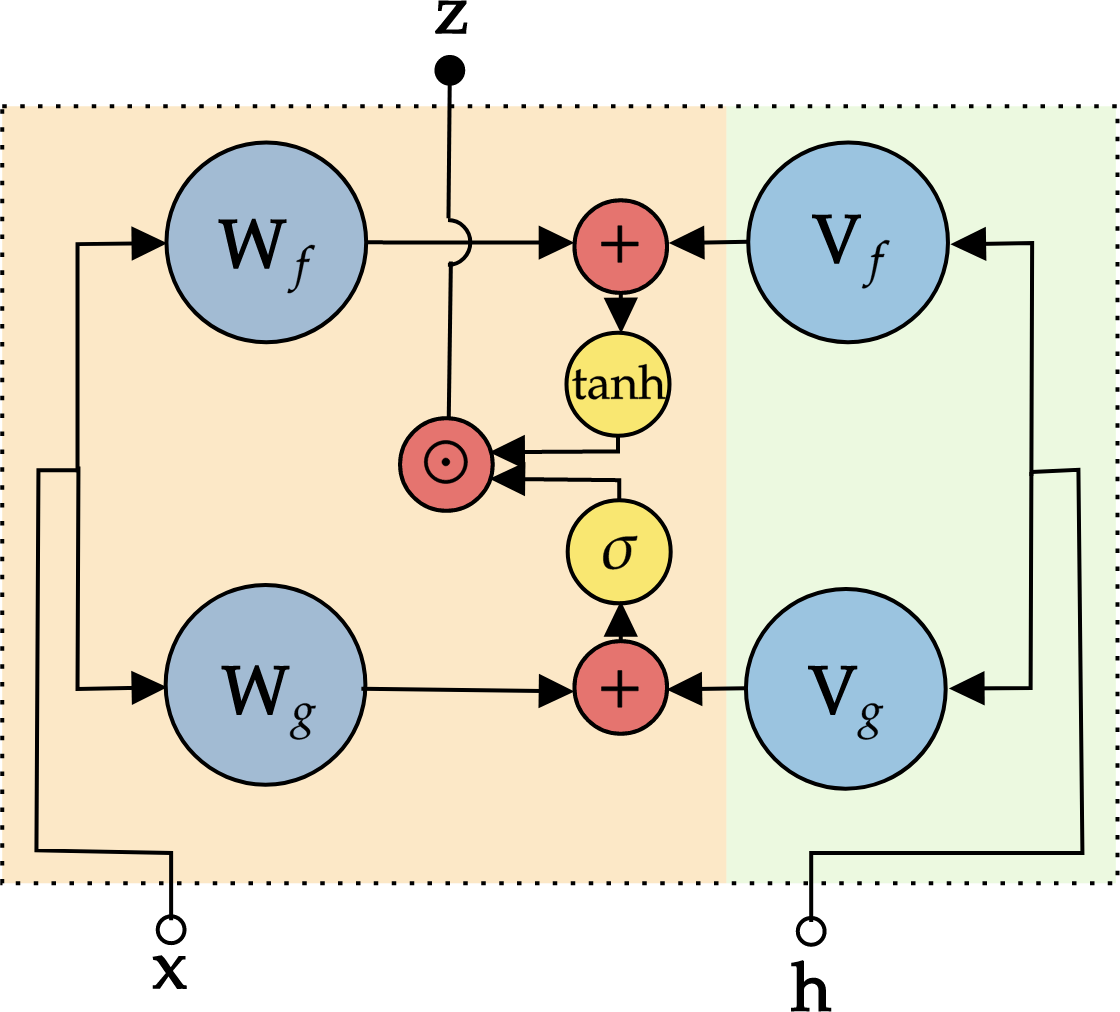}
 \caption{A gated activation unit with conditioning. Input features, $\mathbf{x}$, are processed by a gated activation unit (orange), and the resulting filtering and gating outputs are conditioned (green) based on client one-hot encoding, $\mathbf{h}$, resulting in the output, $\mathbf{z}$. Blue indicates matrix multiplications with learnable weights, yellow are activation functions, and red are element-wise binary operators.}
 \label{fig:CGAU}
\end{figure}

We propose the use of a simple architectural component for enabling a (federated learning) model to identify whether global features are expressed at a client and how each client modulates global patterns.
Specifically, for the classifiers on top of the pre-trained networks, we use a feed-forward neural network with gated activation units and enable the model to condition the units based on the client.
We will refer to units as a conditional gated activation units, CGAU, see \cref{fig:CGAU}.
A federated learning algorithm, such as \textsc{FedAvg}, can then be used to optimize the classifier.

The CGAU consists of two parts.
In a gated part (orange), an input $\mathbf{x}$ is processed by filter- and gate-weights  (learnable $\mathbf{W}_f,\, \mathbf{W}_g\in \mathbb{R}^{D \times N}$, respectively, where $N$ is the number of units and $D$ is the dimensionality of the input) followed by a hyperbolic tangent or a sigmoidal activation function.
The conditioning part (green) shifts the responses of the filter ($\mathbf{W}_f^\top \mathbf{x}$) and gate ($\mathbf{W}_g^\top \mathbf{x}$) before applying the activation functions.
A simplifying view of the process, to provide some intuition, is that the conditioning of the filtering responses modulates the global features (``what the feature is''), and the conditioning of the gating responses controls the client expression of features (``whether this feature is active''), each through corresponding learnable weights $\mathbf{V}_f, \mathbf{V}_g \in \mathbb{R}^{K \times N}$, where $K$ is the number of clients.

In total, the output of the CGAU, $\mathbf{z}$, is:
\begin{equation}
\label{eq:1}
 \mathbf{z} = \overbrace{ \tanh (\mathbf{W}_f^\top \mathbf{x}+\underbrace{\mathbf{V}_f^\top\mathbf{h}}_{\mathrm{\color{green} modulation}} ) }^{\mathrm{\color{orange} filter}} \odot \overbrace{\sigma ( \mathbf{W}_g^\top \mathbf{x} +\underbrace{\mathbf{V}_g^\top\mathbf{h}}_{\mathrm{\color{green}expression}})}^{\mathrm{\color{orange} gate}},
\end{equation}
where $\mathbf{x}$ are and $\mathbf{z}$ are the input to and output of a layer in the classifier (biases omitted, $\odot$ is the Hadamard product, and $\tanh$ and $\sigma$ are the hyperbolic tangent and sigmoidal activation functions).
In the simulated setup, the conditioning is a one-hot encoding of clients IDs, captured as $\mathbf{h}\in\mathbb{R}^{K}$.
Importantly, in a real federated learning setting, we do not need to share the learnt client conditioning; for a particular client, the one-hot encoding, in essence, selects the dimension of the matrices $\mathbf{V_f}$ and $\mathbf{V_g}$ that pertains to the client. This conditioning can simply be maintained locally and can be thought of as an additional local, learnable translation of the filtering and gating responses.
Gated activation units with conditioning in this format have been used both in the WaveNet and PixelCNN architecture \cite{oord2016wavenet, oord2016conditional} in a convolutional form, and are based on previous work on gated activation units in e.g. highway networks \cite{srivastava2015highway} and long short-term memory cells \cite{hochreiter1997long}.

\subsection{Simulating clients with non-IID features} \label{sec:client-simulation}
While non-IID clients, both in terms of their label distribution and feature distributions, are inevitable in realistic federated learning, most data sets that are commonly investigated in deep learning do not have any inherent client identification for the samples of the data set.
One approach to simulate a more realistic scenario in lieu of actual clients is to assign only a subset of all labels to any given client in a set of simulated clients.
This approach mimics a scenario in which the label distributions are non-IID across clients (as in e.g. \cite{Zhao2018}).

However, in realistic settings, we might encounter that clients express a particular class differently, which can also be thought of as having features that are (even class conditionally) non-IID.
In order to simulate such non-IID client feature distributions, we investigate simulated clients where the samples of a particular class are clustered based on the feature embedding from a pre-trained network.

We simulate clients by splitting the data, $\mathcal{D}$, in $K$ nodes, $\mathcal{K} = \{1, \ldots, K \}$.
Each node $k$ has a disjoint data set $\mathcal{D}_k \subset \mathcal{D}$ available for federated training, such that
$\bigcup\limits_{k=1}^{K} \mathcal{D}_{k} = \mathcal{D}$.
For a given data set with $C$ classes, we partition the data in training and test data, and embed the available data using a pre-trained network (cf. \cref{sec:pre-trained}).
We then learn a dimensionality reduction on the training partition of the embedded data through principal component analysis.
For simplicity, we use the first two principal components of the training data and project all the available data onto these.

For each class, we then cluster the training data using a $K$-means clustering, thereby obtaining a set of $C$ times $K$ cluster centroids.
Each of the $K$ clients are then assigned a set of centroids such that samples that are the closest to one of a client's $C$ centroids belongs to that client.
In this manner, we obtain data distributions in the embedding space that are locally clustered and distinct in each client.

\subsection{Quantifying heterogeneity of client feature distributions} \label{sec:frechet}

We are interested in understanding how non-IID data distributions (beyond class distributions) affects federated learning.
We consider data from $K$ clients, where the $k$'th client-specific data $\mathcal{D}_k$ consists of $M_k$ pairs of input/label-pairs $\{\mathbf{x}_m^k, y_m^k\}_{m=0}^{M_k}$.
We can say that the class distribution follows a categorical distribution over $C$ classes, each with some probability for that client, expressible as a vector $\mathbf{p}_k$, such that $y_m^k \sim \mathrm{Cat}(C, \mathbf{p}_k)$. Previous studies have considered how differences in $\mathbf{p}_k$ for different clients affect learning.
In contrast, we are interested in how differences of the input data, $\mathbf{x}_m^k \sim p_k\left(\mathbf{x}\right)$, affect learning. We consider derived features, or embeddings, as learnt in a pre-trained network $\mathbf{z}_m^k = f_\theta(\mathbf{x}_m^k)$ as the basis for investigation---partly inspired by the use of the Fréchet Inception Distance \cite{heusel2017gans} for evaluating the performance of generative adversarial networks.

For the purposes of this paper, we define the overall client data heterogeneity as the average distances from any given client to the rest of the clients in their distributions of embeddings, $\mathbf{z}_m^k \sim p_k\left(\mathbf{z}\right)$.
Under a strong assumption of normality, the level of heterogeneity is quantifiable as a distance between multivariate Gaussians, and we consider the Fréchet distance between two distributions $D_1$ and $D_2$, $d^2(D_1, D_2)$ (among other names also know as the 2\textsuperscript{nd} Wasserstein distance).

The Fréchet distance between two multivariate Gaussians, $D_1=\mathcal{N}\left(\mu_1, \Sigma_1\right)$ and $D_2=\mathcal{N}\left(\mu_2, \Sigma_2\right)$, can be determined as:
\begin{equation}
 d^2(D_1, D_2)=||\mu_1 - \mu_2 ||_2^2 + \Tr \left(\Sigma_1 + \Sigma_2 - 2\left(\Sigma_1 \Sigma_2\right)^{1/2}\right).
 \label{eq:frechet}
\end{equation}

For a given set of clients $\mathcal{K}=\{1, \ldots, K\}$, we are interested in measuring the overall level of heterogeneity, and so we quantify this heterogeneity of clients by determining the distance from any given client to all other clients in their embeddings.
We determine sample mean, $\mu_k$, and covariance, $\Sigma_k$, of the embeddings for both the client-specific data (thus assuming $\mathbf{z}_m^k \sim D_k=\mathcal{N}\left(\mu_k, \Sigma_k\right)$), and a pooling of the data from all other clients ($D_{\mathcal{K} \setminus k}=\mathcal{N}\left(\mu_{\mathcal{K} \setminus k}, \Sigma_{\mathcal{K} \setminus k}\right)$)
We determine the distance using \cref{eq:frechet}, and then use the average distance as our measure of heterogeneity, $\Gamma$:

\begin{equation}
 \Gamma = \frac{1}{K}\sum_{k=1}^{K} d^2(D_k, D_{\mathcal{K} \setminus k})
\end{equation}

\section{Experiments}
\begin{figure*}[tb!]
 \centering
 \begin{subfigure}{0.49\textwidth}
  \centering
  \includegraphics[width=\linewidth]{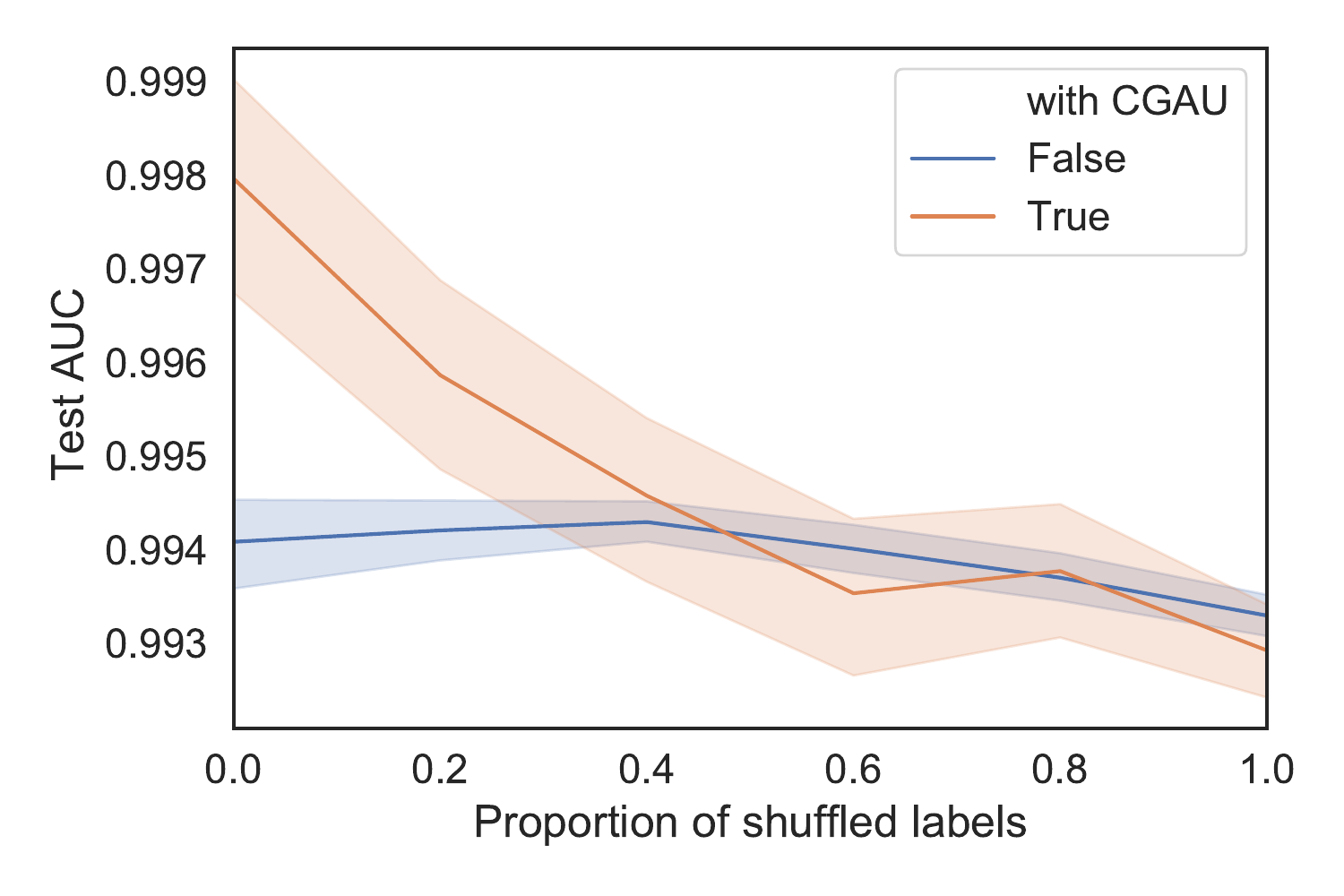}
 \caption{Experiment 1: results on FSD (audio) data. }
 \label{fig:AUDIO}
 \end{subfigure}
 \begin{subfigure}{0.49\textwidth}
  \centering
  \includegraphics[width=\linewidth]{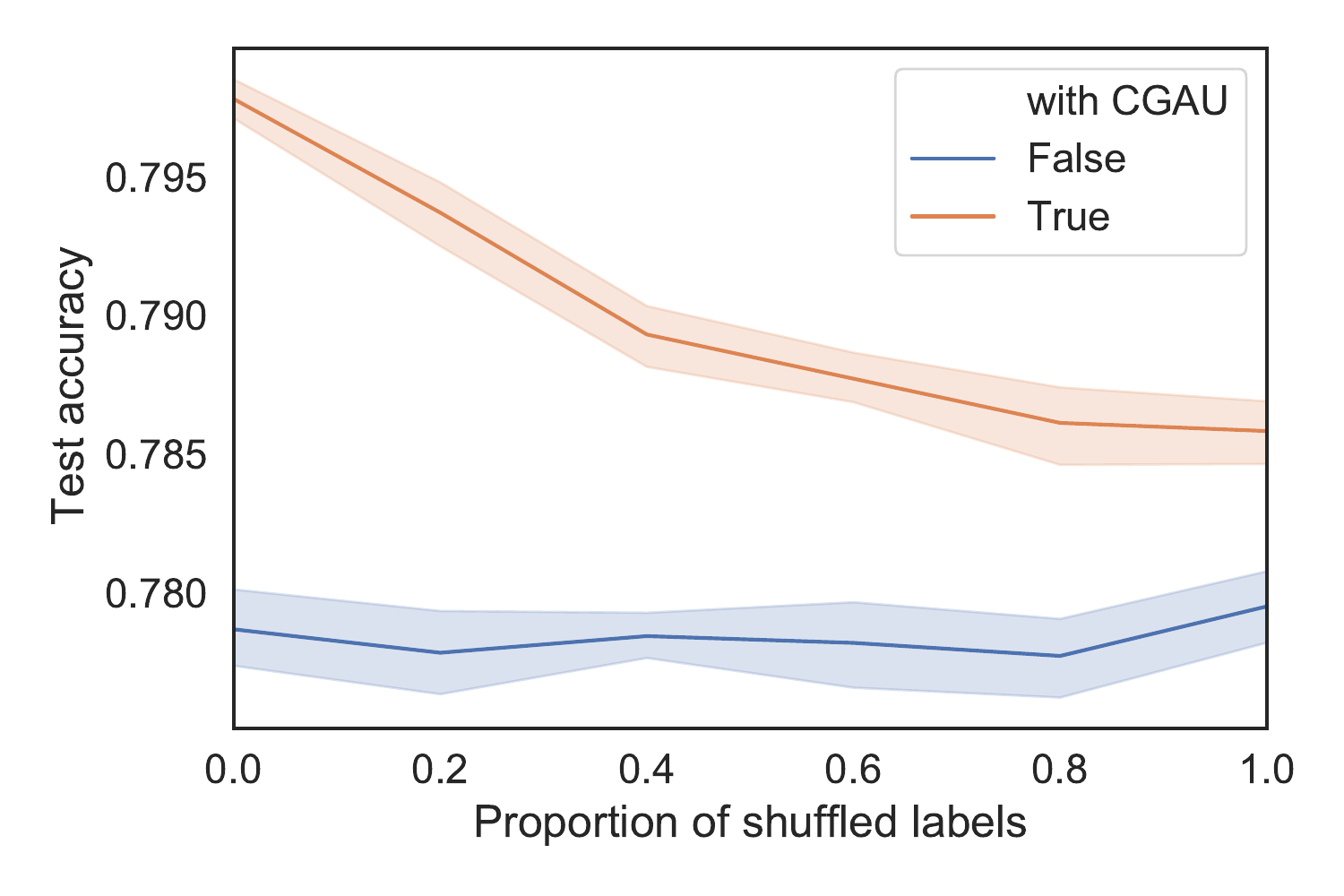}
 \caption{Experiment II: results on CIFAR-10. }
 \label{fig:CIFAR10}
 \end{subfigure}
 \caption{Experimental results. The model trained with CGAU (in red) performs better (or equally well) than the baseline (in blue) at all levels of shuffling. CGAU is particularly beneficial for simulated very heterogeneous (non-IID) clients), marked by the performance difference at no shuffling.}
 \label{fig:results}
\end{figure*}
We use two data sets to evaluate our proposed scheme, one based on audio data and one on image data. We chose the two data sets to cover imbalanced and balanced data as well as binary and multi-class tasks. In both settings we use a pre-trained network that was not trained on the data set at hand. The pre-trained networks are used as feature extractors to provide the features for a classifier. Experiments were carried out with Tensorflow Federated \cite{tensorflow2015whitepaper}.

We investigate the performance of a classifier that utilizes client adaption through conditional gated activation units, as described in \cref{sec:cgau}. We contrast the effect of client adaption with a standard feed-forward neural network with rectified linear units without conditioning.

We simulated clients with non-IID features by clustering the embedding features in the manner described in \cref{sec:client-simulation}.
We can control the level of client heterogeneity by shuffling a certain percentage of the sample client assignments. For a shuffling proportion of 0.0, no samples are randomly re-assigned to any of the $K$ clients. In this case, we consider the simulated clients to be maximally non-IID (all samples that are the closest to a particular centroid are collected in one client). On the other hand, a shuffling of 1.0 corresponds to completely random assignment of samples to clients.

In Experiment I, we investigate data from the Freesound Database (FSD) Kaggle 2018 data set \cite{fonseca2018general}. FSD contains approximately 11k audio clips from 41 different classes (such as laughter, keys jangling, writing, trumpet, and coughing). In Experiment II, we investigate CIFAR-10 \cite{krizhevsky2009learning}, a popular image dataset featuring 32x32 resolution natural images from ten different classes.  Additionally, we illustrate how the conditioning through CGAU changing the neural network solution on a XOR-problem in \cref{sec:toydata}.

\subsection{Experiment I: audio cough detection}
In our investigation of audio data, we construct a binary problem from the FSD by subdividing the labels into a positive class of audio labelled as cough, and a much larger negative class of any other label in FSD.
FSD as a whole has a total of 273 examples of the cough label.

We use a pre-trained MobileNet \cite{howard2017mobilenets} called YamNet\footnote{Maintained by M. Plakal and D. Ellis and available in the \href{github.com/tensorflow/models/tree/master/research/audioset/yamnet}{Tensorflow AudioSet research repository}.} trained on the AudioSet data \cite{gemmeke2017audio} to obtain embeddings. Audio inputs are resampled to 16 kHz mono signals, and converted to stabilized log-mel spectrograms. YamNet outputs a 1024-dimensional vector for each patch in the log-mel spectrogram, where a patch corresponds to 960 ms of waveform input (only full 960 ms patches are considered and any remainder of the signal is dropped). We obtain an embedding robust to varying time-length of the FSD audio samples by max-pooling the patch-features across the patches (the temporal dimension).

We train a 2-layer classifier with 64 units with 50 \% dropout between layers.
The classifier is trained using \textsc{FedAvg} \cite{mcmahan2016communication} with 10 simulated clients, and at each round of federated learning all 10 clients were included.
Each client completed 10 steps of gradient descent with batch sizes of 32 samples, or until all client data had been seen once---each client had a variable size of data set, seeing as samples are assigned based on proximity to the cluster centroids.
The clients utilized a stochastic gradient descent optimizer\footnote{A momentum and a decay was specified, but at a later stage it was discovered that the states were erroneously resat at each round (i.e. at each 10 steps), effectively thus having no decay, and momenta building over only 10 steps.}.
We retain the original FSD training and test partitions.
While training, we monitor the loss of a held-out sub-partition of 5 \% the training set (a validation partition) by centrally collecting the outcomes on the validation data points at each client.
The best model (model weights) with the lowest cross-entropy loss on the validation partition across clients after a total of a 1000 rounds of federated learning is then used in the final evaluation on the test set.

We determine the average Fréchet distances from any given client's features to all others (as described in \cref{sec:frechet}). The results averaged across replicates of the shuffling proportion are shown in \cref{tab:frechet}.

We evaluate the performance for shuffling proportions ranging from 0.0 to 1.0 in steps of 0.2 with 12 repetitions of the experiment at each level for both with and without client adaption. Since the data is imbalanced, we measure the model performance in terms of the area under the receiver operating characteristic (AUC).
The results are shown in \cref{fig:AUDIO}.
We see that a model without client adaption (in blue) performs consistently at about 0.994 across the range of shuffling.
The model with client adaption (orange) outperforms this baseline model for the most non-IID clients (lowest levels of shuffling), and has an AUC of about 0.998, thus improving performance for the most heterogeneous simulated clients.
For more homogeneous clients, we see no discernible difference between the performance of the two models.
\begin{table*}
\centering
\caption{Client average Fréchet distances, $\Gamma$, in relation to proportion of shuffled labels, for the audio data in Experiment 1. Increasing the proportion of shuffled labels decreases the distances.}
\label{tab:frechet}
\begin{tabular}{lrrrrrr}
\toprule
Proportion of shuffled labels &  0.0 &  0.2 &  0.4 &  0.6 & 0.8 & 1.0 \\
\midrule
Average client Fréchet distance & 1784.4 & 980.0 & 550.5 & 271.0 & 91.8 & 19.9 \\
\bottomrule
\end{tabular}
\end{table*}

\subsection{Experiment II: image label classification}

To show that our approach also works on a balanced and more challenging data set, we show results on CIFAR-10 \citep{krizhevsky2009learning}.

We use an Inception architecture pre-trained\footnote{Model from \href{https://github.com/pytorch/vision/tree/master/torchvision}{github.com/pytorch/vision/tree/master/torchvision}} on ImageNet
 \cite{szegedy2016rethinking,deng2009imagenet}.
The embeddings were extracted from the next to last layer.
To align the CIFAR-10 images with the resolution of ImageNet, we upsampled the images to 299x299 pixels.
We randomly partitioned the training data set into 80 \% training data and 20 \% validation data.
The classifier network mirrors the one used in the audio-experiment, except the dimensions were increased to 128 hidden units each in the two layers.
The federated learning scheme is similar to the audio experiment, but utilizes an increased learning rate of 0.01, and all models were trained for 500 epochs.
All configurations were run five times with different seeds. Training with CGAU took roughly 85\% longer per epoch in the simulated setting.

We show results for this task in \cref{fig:CIFAR10}. Since the class sizes of CIFAR-10 are balanced, we show model performance in terms of the accuracy.
The model with client adaption outperforms the baseline (without client adaption) at all levels of label shuffling. However, it works particularly well when there is a large difference in the training distribution between clients.

\section{Discussion}
In \cref{fig:AUDIO} and \cref{fig:CIFAR10} we see that the models trained with the proposed conditional gated activation unit outperforms or perform equal to the baseline model at all levels of diversity between clients. In additional experiments on a toy dataset we show that CGAU does enable the network to learn client specific features and feature expressions (results in \cref{sec:toydata}).

Notably, CGAU is particularly useful when there is a high diversity between clients (no or little label shuffling).
This may be more pronounced due to the effects of transfer learning, i.e. fine-tuning an already pre-trained model on the available task-specific data.
As a result, the embeddings from the lower layers may be worse at  ``discounting''' client-variant features, making it particularly important to learn client-specific embeddings along with global embeddings.

In preliminary experiments on CIFAR-10, we found that the gain in using CGAUs were not as pronounced when the feature embeddings are extracted from a network pre-trained with CIFAR-10---and not with ImageNet as shown on \cref{fig:CIFAR10}. We theorize that CGAUs are particularly useful when the feature extractor was not trained on the same data distribution, as the feature extractor will not be as adept at extracting invariant features while discarding spurious features in the embedding.

To ensure that low shuffling does correspond to higher diversity in between client data distributions we provide the average Fréchet distance from each individual client to the remaining data, $\Gamma$, in \cref{tab:frechet}.
We see that label shuffling rates correlate well with $\Gamma$, implying that clustering based on the embedding may be a good alternative, or addition, to class clustering for artificially creating non-IID data sets.
In particular it allows us to measure the impact of data that is non-IID even when conditioned on the classes.

A surprising effect is that applying CGAU results in predictive accuracy increasing for diverse data distributions (low proportion of shuffling). We would expect the accuracy to be decreased for the base model for diverse data distributions as the models diverge between synchronizations and to stay relatively equal for the model with CGAU. Instead the base model performs relatively equal across all levels of label shuffling whereas the CGAU model performs better for a low proportion of label shuffling. We hypothesize that this is due to the model learning the specific data distribution from each client.

While the cough detection problem is highly imbalanced, the classification task is a less challenging problem when using a well-suited pre-trained network.
A considerable portion of the audio samples are, e.g., musical instruments with tonal qualities, that are quite straightforward to distinguish from coughs.
This is also evident from the experimental results, where even the baseline effectively solves the task with an AUC of about 0.994.

A marked difference in problem complexity between Experiment I and II is evident in the performance difference at homogeneous clients (shuffling of 1.0). The model capacity (if naïvely measured as parameter count) is doubled by the filtering and gating alone in using the CGAU compared to the baseline. This increase in capacity is not beneficial for the homogeneous clients in the audio task, but does increase the test accuracy on the CIFAR-10 problem from about 0.780 to 0.785. This also increases the computational power needed for each round of learning, potentially exacerbating problems with e.g. stragglers in real federated learning systems; whether the performance gains of CGAU is worth such trade-offs remains to be investigated.

In our experiments, we worked with a relatively low number of clients (ten clients for both experiments). This was done to ensure that the clustering as described in \cref{sec:frechet} resulted in semantically meaningful clusters of data samples. Examining the effect with a larger number of clients is needed, seeing as FL usually works with a (much) larger number of clients.
A potential alternative solution may be to use a clustering scheme as suggested in \citet{Ghosh2019} to find clients with shared attributes. However, sharing characteristics specific to a client in a privacy-compliant manner may be a challenge. %

Incorporating client-information directly in federated learning models, even if only locally, is a potential opening for attacks on privacy. The client-specific conditioning does not need to be shared, yet any use of conditioning in the manner of the investigated CGAU would need to be evaluated for robustness to attacks.

\section{Conclusion}
Extant previous work has shown that models trained in a federated learning manner converge much slower if the used data sets are non-IID between clients \cite{Li2019}. Since this is an extremely common characteristic when learning from sensitive user data, it presents a serious hindrance to the utilization of federated learning.

We present a simple approach to reduce the impact of local features by learning patterns specific to each client along with the global model. In experiments we show that our approach outperforms the baseline for scenarios with heterogeneous clients. We find evidence that our approach may be particularly beneficial when using a transfer learning approach to extract embeddings.

\clearpage
\bibliography{references}
\bibliographystyle{icml2020}
\clearpage
\appendix
\section{CGAU on XOR dataset}
\label{sec:toydata}
\begin{figure*}
\begin{subfigure}{.5\textwidth}
\centering
\includegraphics[width=.8\linewidth]{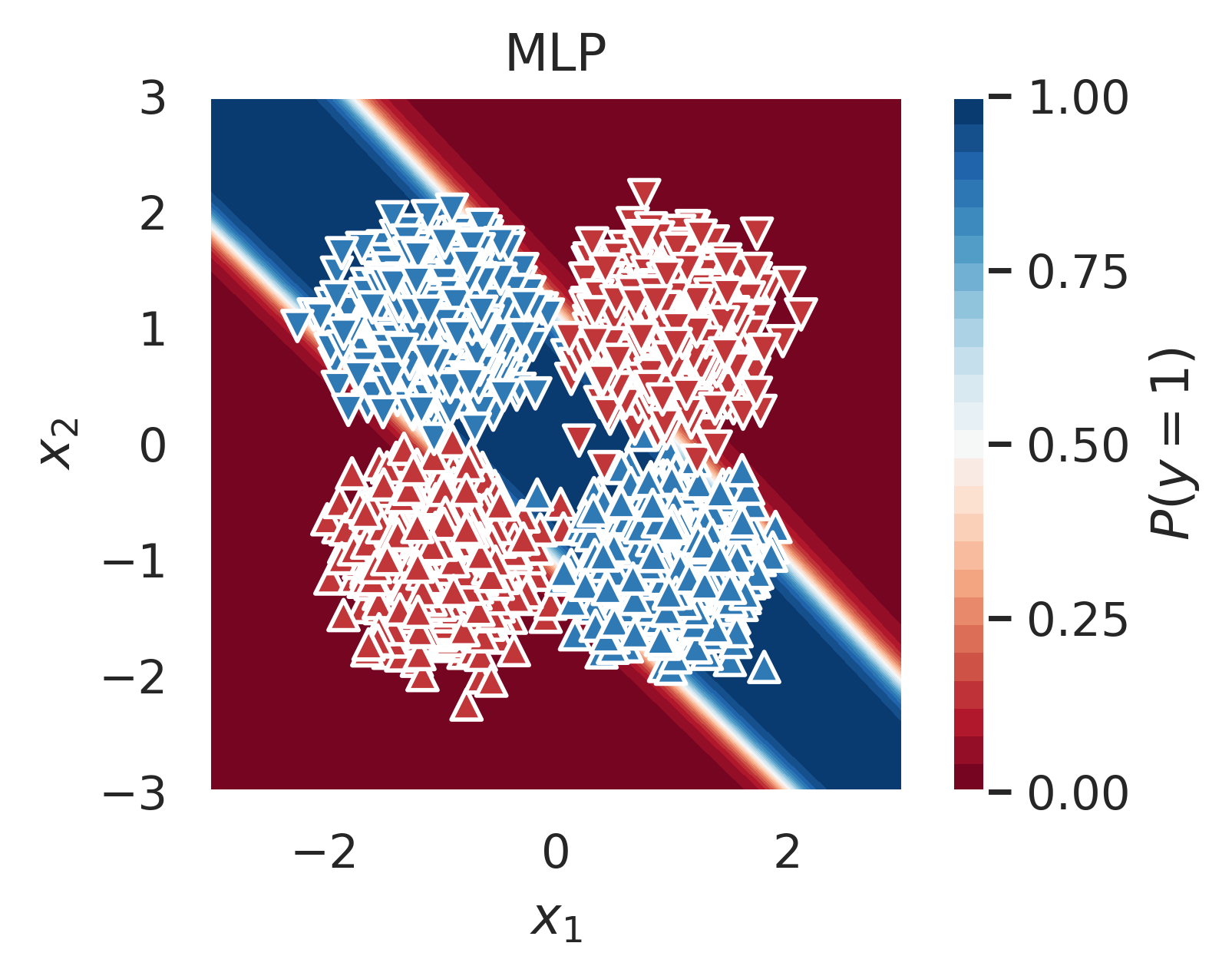}
\caption{Decision boundaries for MLP.}
\label{fig:xor-mlp}
\end{subfigure}
\hfill
\begin{subfigure}{.5\textwidth}
\centering
\includegraphics[width=.8\linewidth]{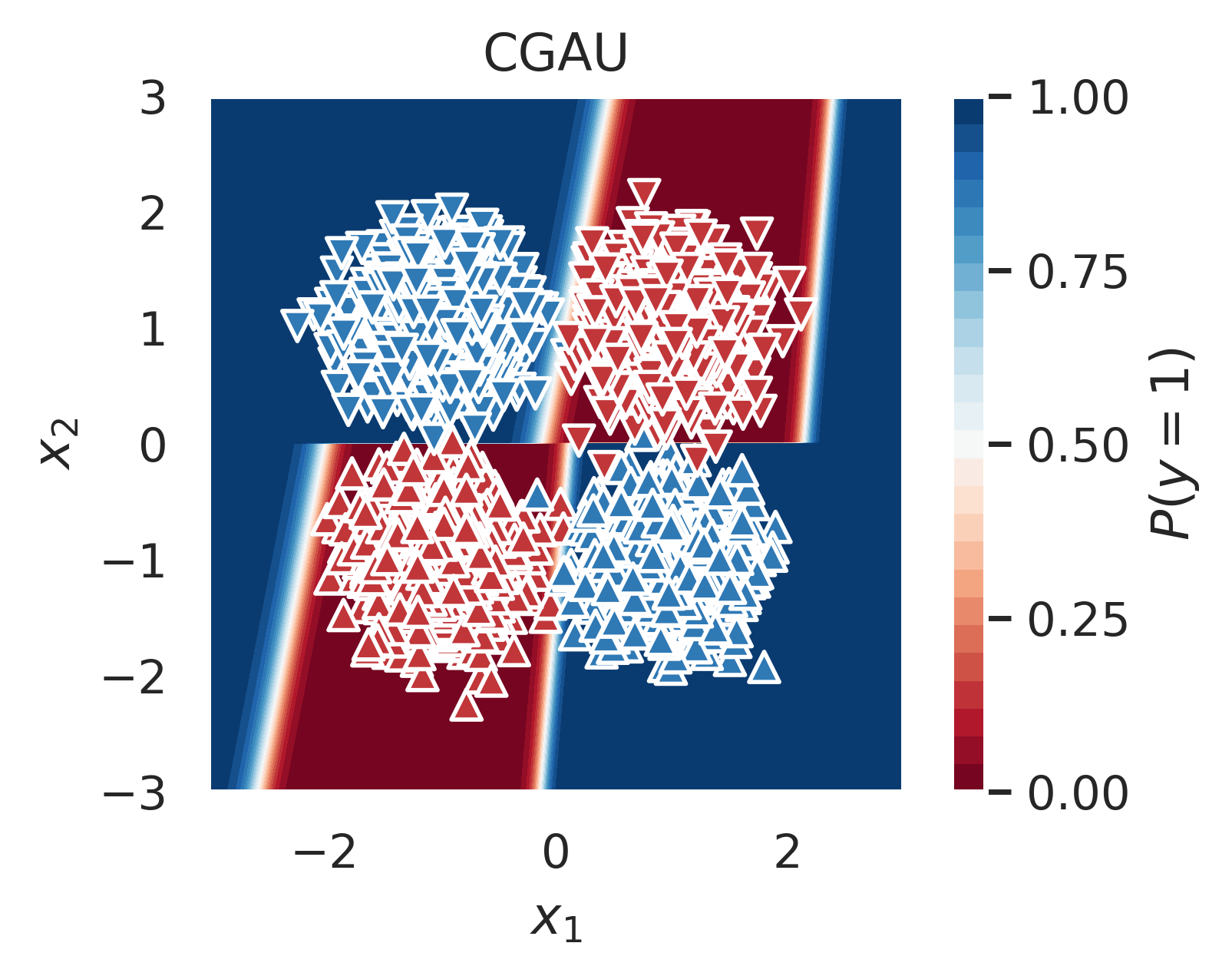}
\caption{Decision boundaries for CGAU.}
\label{fig:xor-cgau}
\end{subfigure}
\hfill
\begin{subfigure}{.5\textwidth}
\centering
\includegraphics[width=.8\linewidth]{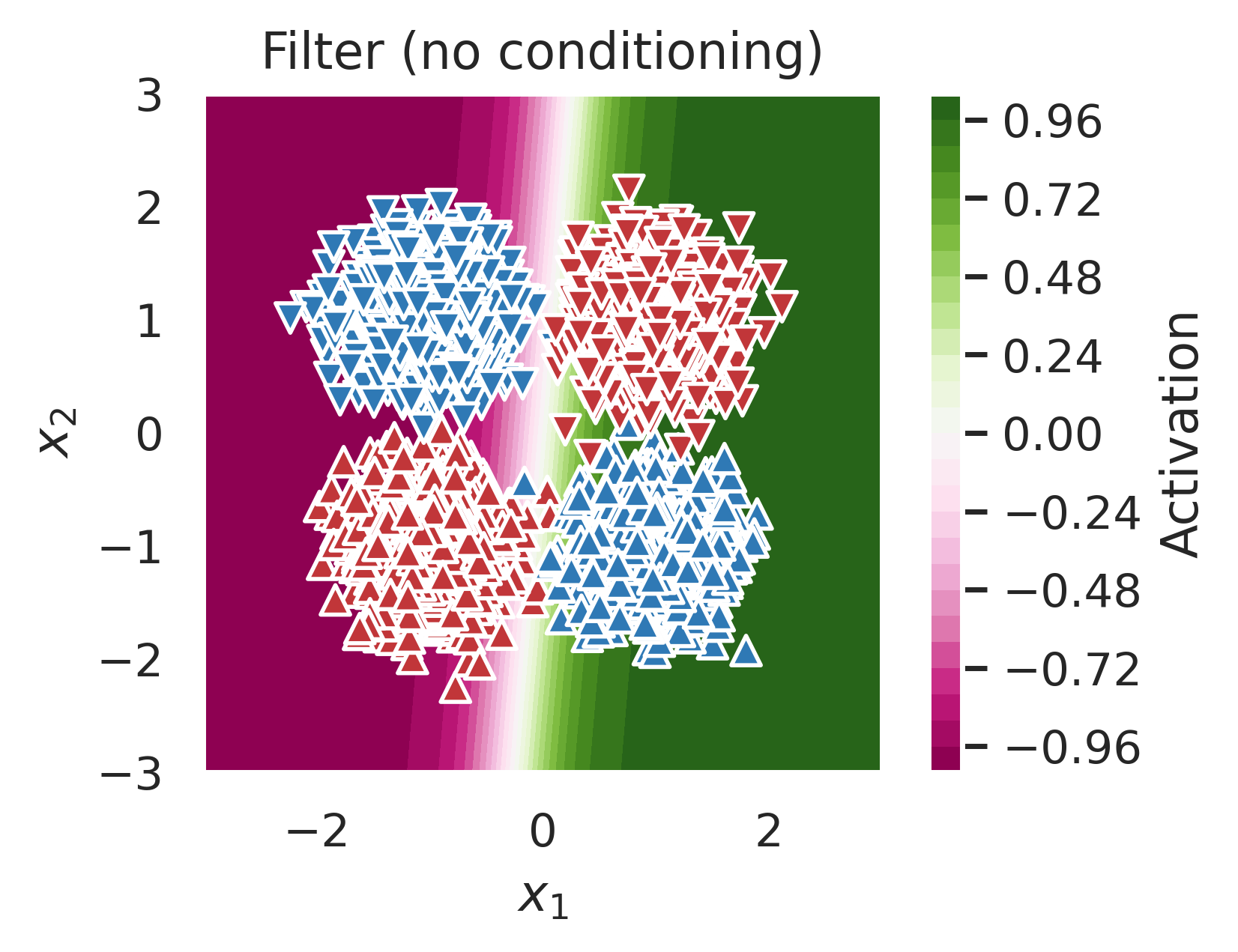}
\caption{CGAU filter activation without conditioning.}
\label{fig:xor-cgau-filter-uncond}
\end{subfigure}
\begin{subfigure}{.5\textwidth}
\centering
\includegraphics[width=.8\linewidth]{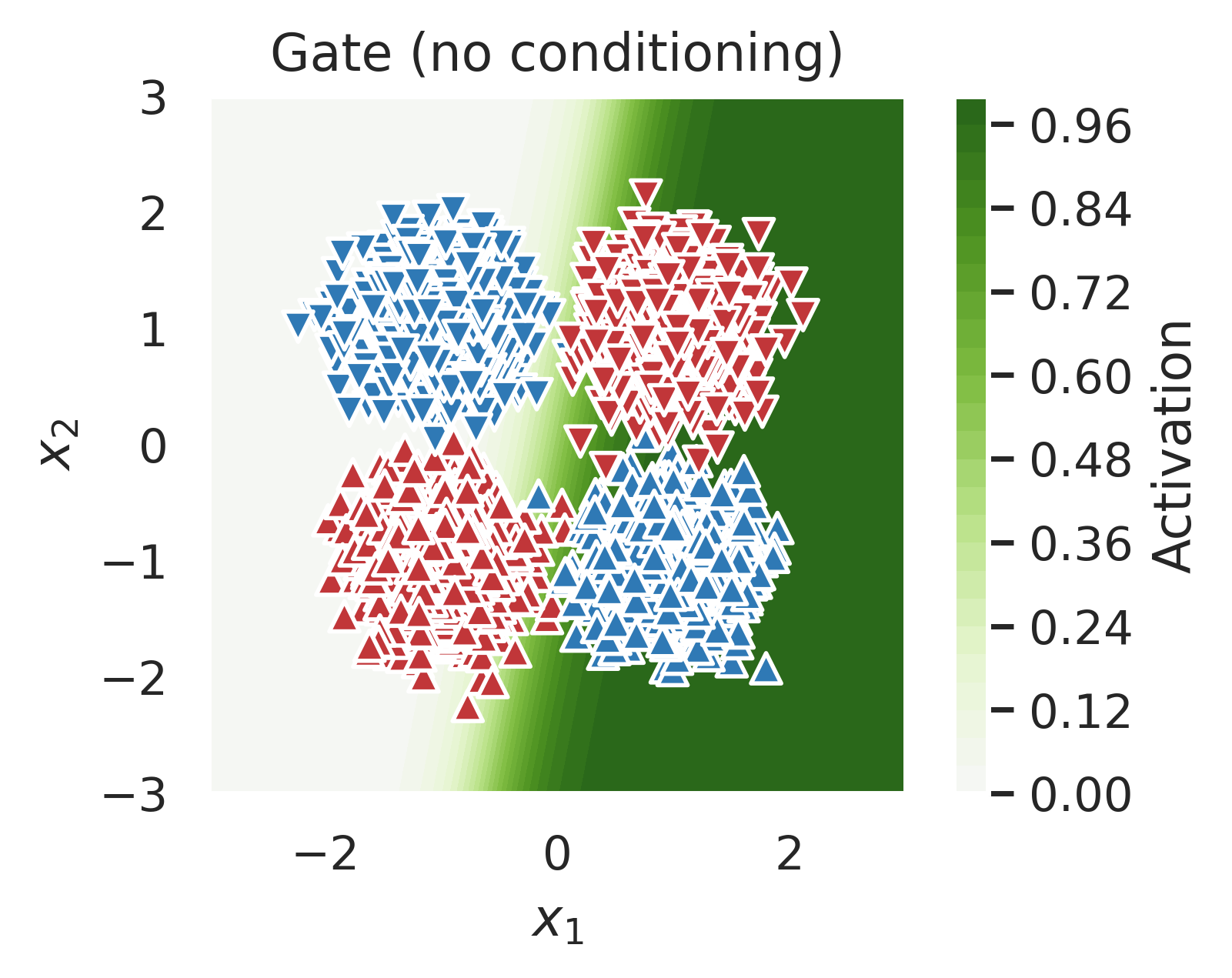}
\caption{CGAU gating activation without conditioning.}
\label{fig:xor-cgau-gate-uncond}
\end{subfigure}

\begin{subfigure}{.5\textwidth}
\centering
\includegraphics[width=.8\linewidth]{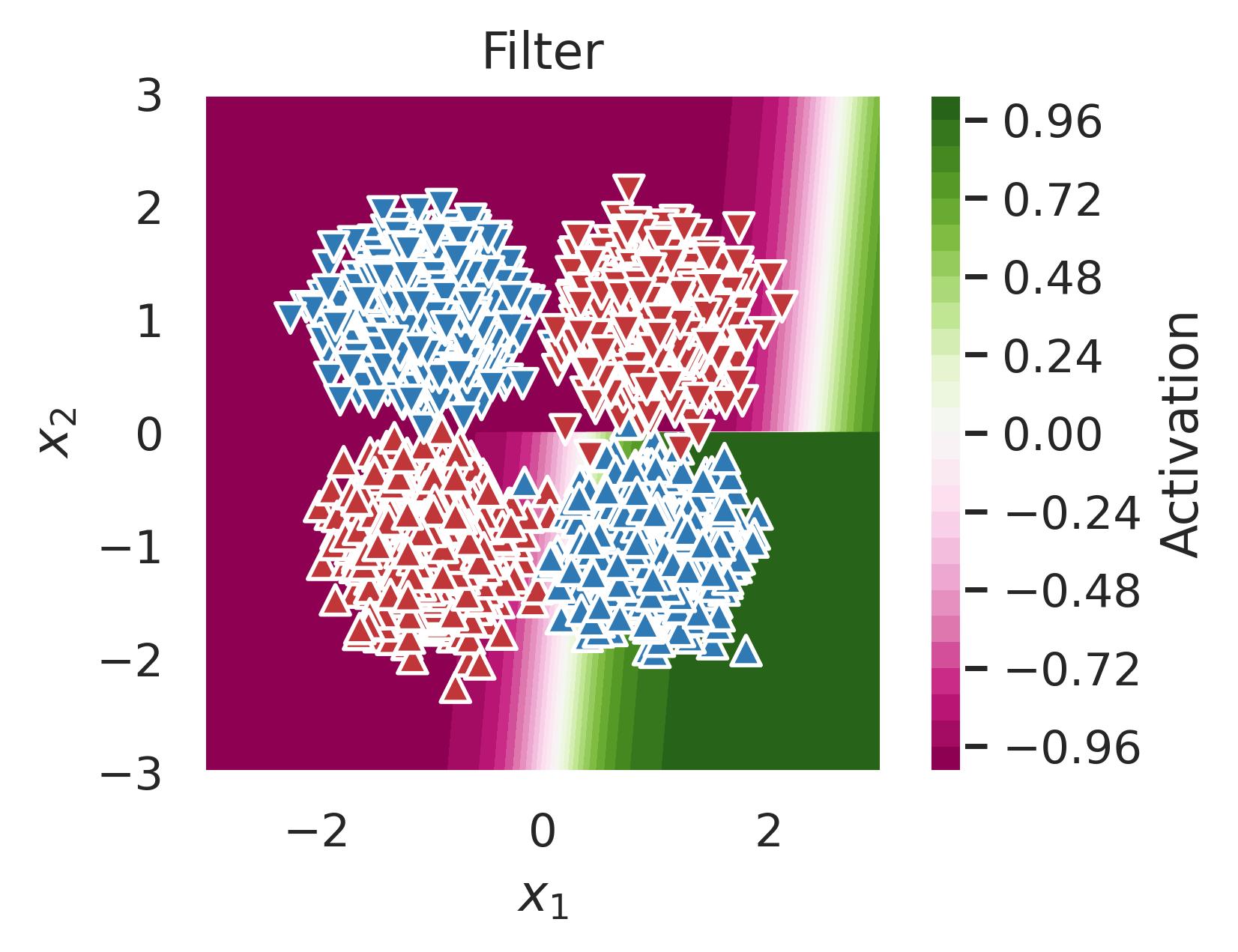}
\caption{CGAU filter activation with conditioning.}
\label{fig:xor-cgau-filter}
\end{subfigure}
\hfill
\begin{subfigure}{.5\textwidth}
\centering
\includegraphics[width=.8\linewidth]{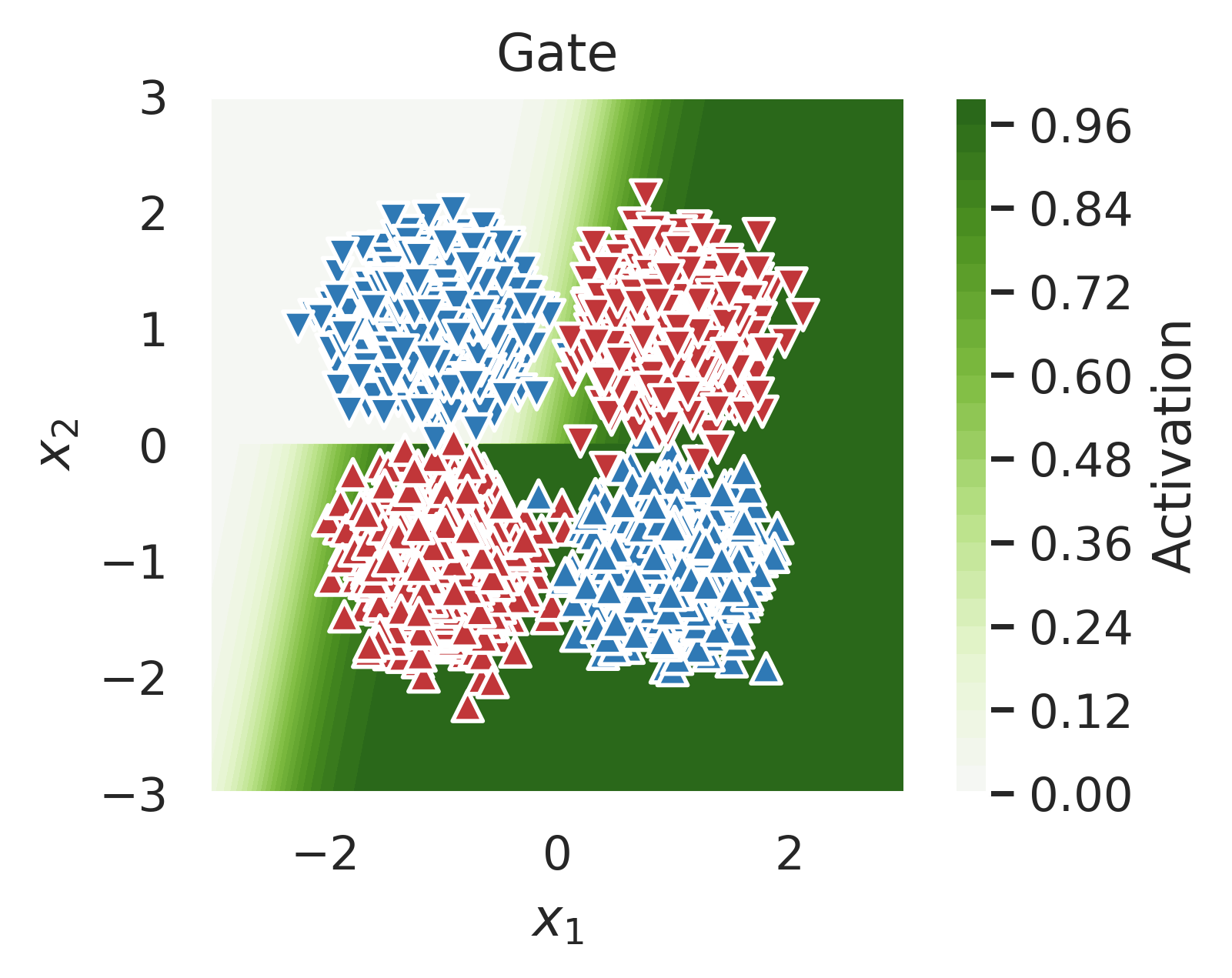}
\caption{CGAU gate activation with conditioning.}
\label{fig:xor-cgau-gate}
\end{subfigure}
\hfill
\begin{subfigure}{.5\textwidth}
\centering
\includegraphics[width=.8\linewidth]{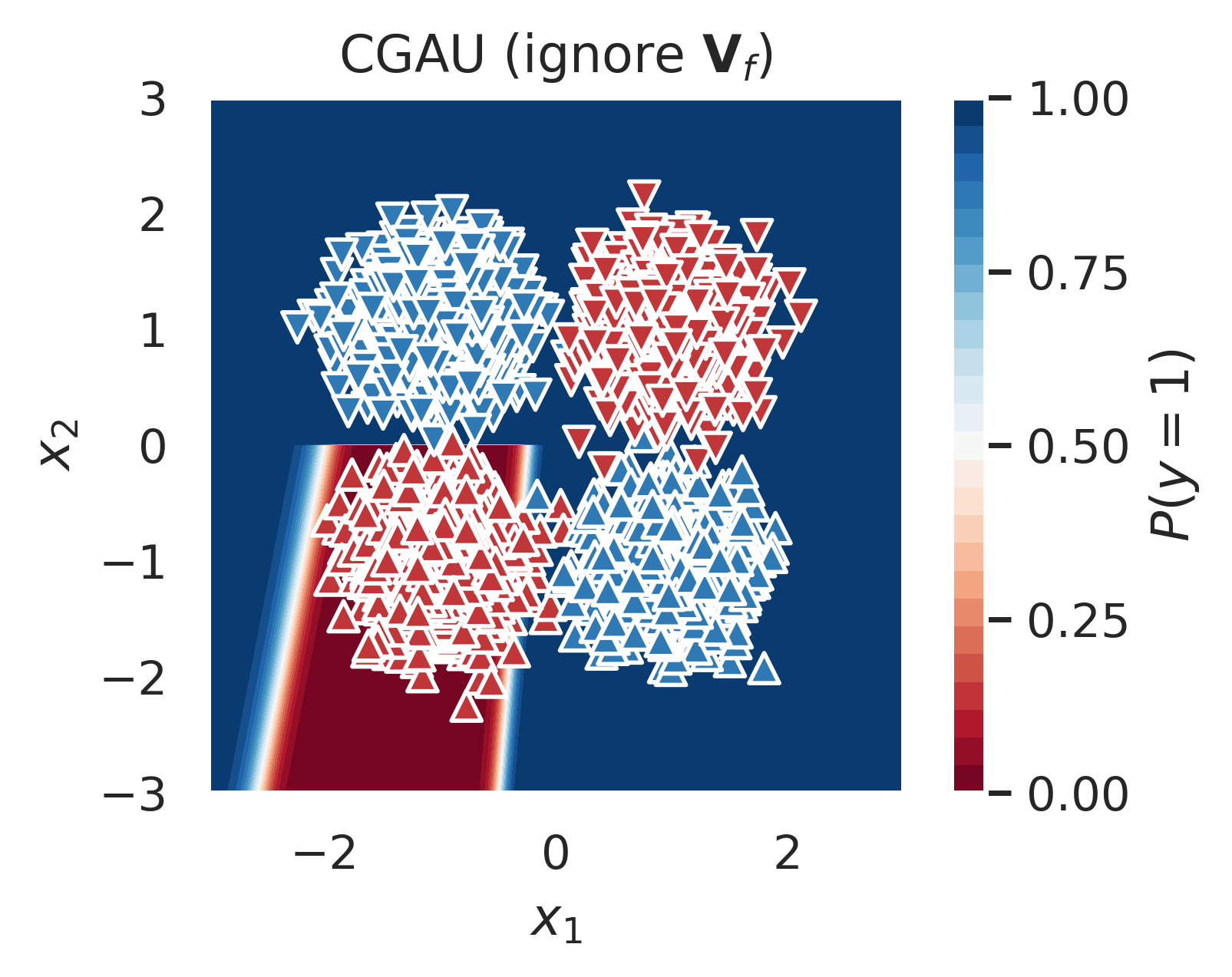}
\caption{Decision boundaries for CGAU without client modulation.}
\label{fig:xor-cgau-ignorevf}
\end{subfigure}
\begin{subfigure}{.5\textwidth}
\centering
\includegraphics[width=.8\linewidth]{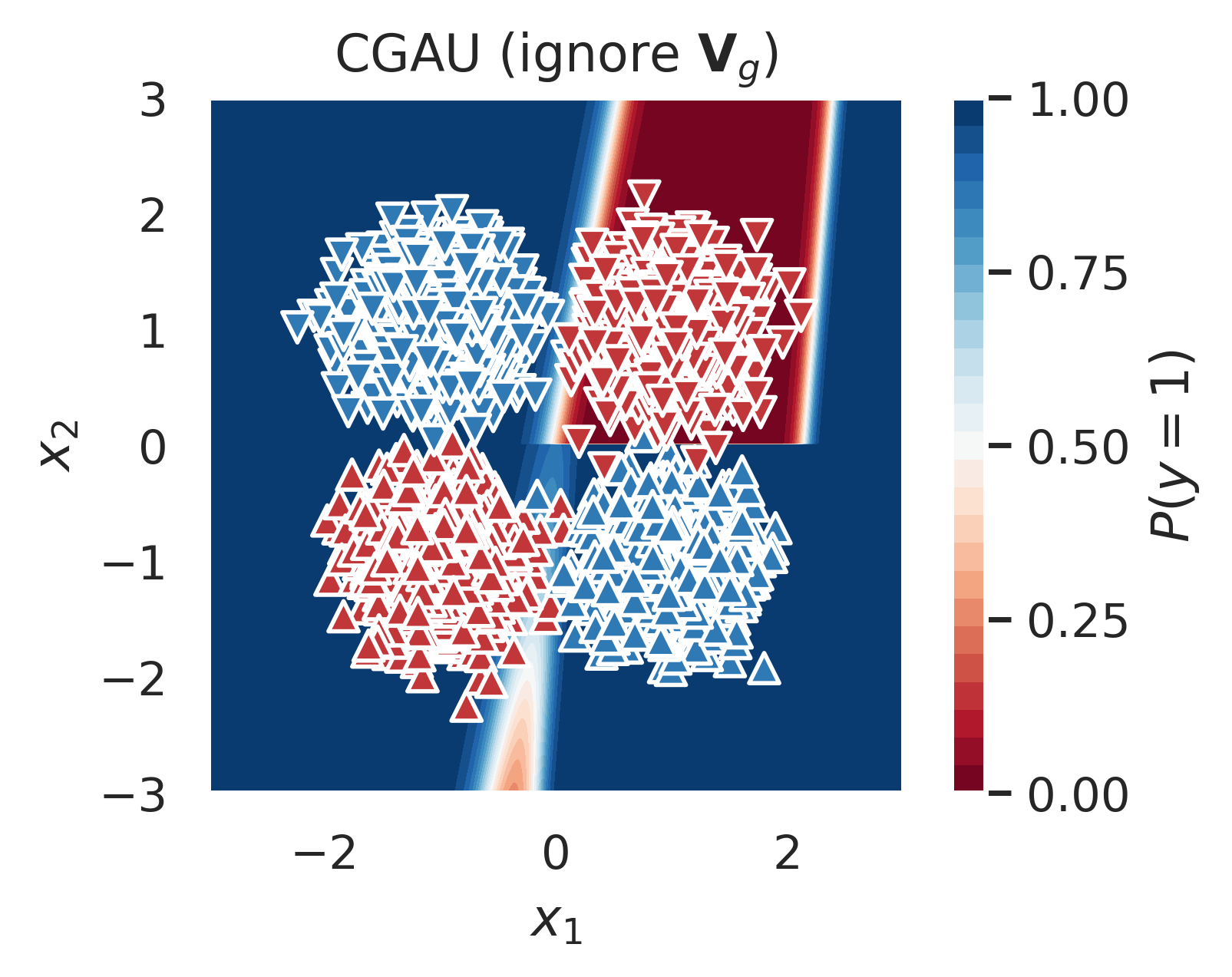}
\caption{Decision boundaries for CGAU with client expression.}
\label{fig:xor-cgau-ignorevg}
\end{subfigure}

\label{fig:xor}
\end{figure*}

We can illustrate the workings of the CGAU by investigating an augmented version of the classic XOR-problem.
We consider a two-class problem with two features, $x_1$ and $x_2$. The negative class (red) consists of two clusters for which the features have the same sign, and the positive class (blue) is characterized by having the features of opposite sign. The augmentation to the XOR-problem is that we consider the clusters to be from two different clients: client one (``up client'') has only positive $x_2$, and client two has only negative $x_2$ (``down client''). Ignoring the clients and solving the problem using a multi-layer perceptron (MLP, with two hidden units) results in a solution of the form shown on \cref{fig:xor-mlp}, whereas a CGAU with one gated unit ($N=1$) results in a solution of the form shown on \cref{fig:xor-cgau}.

The un-conditioned filter and gate outputs are shown on \cref{fig:xor-cgau-filter-uncond} and \cref{fig:xor-cgau-gate-uncond}.
We see how the filter has learnt a split on the sign of $x_1$, and the gate is shutting off any information for negative $x_1$.
Looking at \cref{fig:xor-cgau-filter} and \cref{fig:xor-cgau-gate}, we see how the CGAU has learnt to modulate the up clients filter feature by moving the shift in activation towards more positive $x_1$, and similarly, we see that the it has learnt to shift the gating for the down client towards more negative $x_1$.

In a sense, this enables the CGAU to solve the problem for the down client using ``client-specific expression'', and enables the CGAU to solve the problem for the up client by using ``client-specific modulation''; this assertion can be confirmed by turning off the two types of conditioning in making a plot like \cref{fig:xor-cgau}, which is shown on \cref{fig:xor-cgau-ignorevg} and \cref{fig:xor-cgau-ignorevf}.

\end{document}